\documentclass[10pt,twocolumn,letterpaper]{article}

\usepackage{cvpr}
\usepackage{times}
\usepackage{epsfig}
\usepackage{graphicx}
\usepackage{amsmath}
\usepackage{amssymb}
\usepackage{color}
\usepackage{microtype}
\usepackage{algorithm}
\usepackage{algorithmic}
\usepackage{ifthen}
\usepackage{subcaption}
\usepackage{wrapfig}
\usepackage{extarrows}
\usepackage{booktabs}
\usepackage{stackengine}
\usepackage[breaklinks=true,colorlinks,bookmarks=false]{hyperref}

\definecolor{ChongyangColor}{rgb}{0.7,0,0}
\definecolor{MengleiColor}{rgb}{0.7,0,0.7}
\newcommand{\chongyang}[1]{{\color{ChongyangColor} Chongyang: #1}} 
\newcommand{\menglei}[1]{{\color{MengleiColor} Menglei: #1}} 
\newcommand{\snam}[1]{{\color{blue} Seonghyeon: #1}} 
\newcommand{\ning}[1]{{\color{blue} Ning: #1}} 
\newcommand{\SJ}[1]{{\color{green} SJ: #1}} 
\newcommand{\note}[1]{{\it\color{blue} #1}}
\newcommand{\commentout}[1]{} 

\renewcommand{\chongyang}[1]{}
\renewcommand{\snam}[1]{}
\renewcommand{\ning}[1]{}
\renewcommand{\menglei}[1]{}
\renewcommand{\SJ}[1]{}
\renewcommand{\note}[1]{}

\newcommand{\eref}[1]{Eq.~(\ref{#1})}

\newcommand{\Fref}[1]{Figure~\ref{#1}}

\newcommand{\Sref}[1]{Section~\ref{#1}}
\newcommand{\Aref}[1]{Algorithm~\ref{#1}}

\newcommand\figurevspace{-12pt} 

\newcommand{\amos}{\mathcal{A}}
\newcommand{\tlvdb}{\mathcal{B}}
\newcommand{\timestamp}{t}
\newcommand{\image}{\mathbf{I}}
\newcommand{\inputimage}{\image}
\newcommand{\outputimage}{\hat{\image}}
\newcommand{\video}{\mathbf{V}}

\newcommand{\latentz}{\mathbf{z}}
\newcommand{\encoder}{\mathbf{E}}
\newcommand{\decoder}{\mathbf{D}}

\newcommand{\generator}{\mathcal{G}}
\newcommand{\generatoramos}{\mathcal{G}_{\amos}}
\newcommand{\generatortlvdb}{\mathcal{G}_{\tlvdb}}
\newcommand{\generatorencoder}{\encoder_{\generator}} 
\newcommand{\generatordecoder}{\decoder_{\generator}} 

\newcommand{\discriminator}{\mathcal{D}}
\newcommand{\conditionaldiscriminator}{\discriminator_{c}}
\newcommand{\unconditionaldiscriminator}{\discriminator_{u}}
\newcommand{\discriminatorencoder}{\encoder_{\discriminator}} 
\newcommand{\discriminatoramos}{\mathcal{D}_{\amos}}
\newcommand{\discriminatortlvdb}{\mathcal{D}_{\tlvdb}}

\newcommand{\frameset}{\mathbf{S}}
\newcommand{\amosframeset}{\frameset_{\amos}} 
\newcommand{\fakeamosframeset}{\hat{\frameset}_{\amos}} 
\newcommand{\fakeamospairset}{\tilde{\frameset}_{\amos}} 
\newcommand{\tlvdbframeset}{\frameset_{\tlvdb}} 

\newcommand{\loss}{l}
\newcommand{\unconditionalloss}{\loss_{u}}
\newcommand{\unconditionallossamos}{\loss_{u, \amos}}
\newcommand{\unconditionallosstlvdb}{\loss_{u, \tlvdb}}
\newcommand{\conditionalloss}{\loss_{c}}

\newcommand{\reconstructionloss}{\loss_{r}}
\newcommand{\objectivefunc}{\mathbb{E}}
\newcommand{\normaldistribution}{\mathcal{N}}

\newcommand{\learningrate}{\eta}
\newcommand{\networkparams}{\theta}

\newcommand{\energyfunc}{\mathcal{E}}
\newcommand{\pixel}{p}
\newcommand{\anotherpixel}{q}
\newcommand{\neighborhood}{\mathbf{N}}
\newcommand{\pixelweight}{w}
\newcommand{\finalimage}{\bar{\image}}
\newcommand{\smoothnessweight}{\mu}
\newcommand{\lineartransform}{\mathbf{T}}
\newcommand{\scalingfactor}{s}
\newcommand{\bias}{b}

\cvprfinalcopy 

\def\cvprPaperID{0673} 

\graphicspath{
{images/}
{images/results_teaser/}
}

\ifcvprfinal\fi
\begin{document}

\title{End-to-End Time-Lapse Video Synthesis from a Single Outdoor Image}

\author{Seonghyeon Nam\textsuperscript{1}\hspace{0.3in}
Chongyang Ma\textsuperscript{2}\hspace{0.3in}
Menglei Chai\textsuperscript{2}
\vspace{2pt}\\
William Brendel\textsuperscript{2}\hspace{0.3in}
Ning Xu\textsuperscript{3}\hspace{0.3in}
Seon Joo Kim\textsuperscript{1}
\vspace{4pt}
\\
\textsuperscript{1}{Yonsei University}\hspace{0.3in} \textsuperscript{2}{Snap Inc.}  \hspace{0.3in} \textsuperscript{3}{Amazon Go}
}

\maketitle
\ifcvprfinal\thispagestyle{empty}\fi

\begin{abstract}
Time-lapse videos usually contain visually appealing content but are often difficult and costly to create. In this paper, we present an end-to-end solution to synthesize a time-lapse video from a single outdoor image using deep neural networks. Our key idea is to train a conditional generative adversarial network based on existing datasets of time-lapse videos and image sequences. We propose a multi-frame joint conditional generation framework to effectively learn the correlation between the illumination change of an outdoor scene and the time of the day. We further present a multi-domain training scheme for robust training of our generative models from two datasets with different distributions and missing timestamp labels. Compared to alternative time-lapse video synthesis algorithms, our method uses the timestamp as the control variable and does not require a reference video to guide the synthesis of the final output. We conduct ablation studies to validate our algorithm and compare with state-of-the-art techniques both qualitatively and quantitatively.
\end{abstract}

\section{Introduction}
\label{sec:intro}

Time-lapse videos are typically created by using a fixed or slowly moving camera to capture an outdoor scene at a large frame interval.
This unique kind of videos is visually appealing since it often presents drastic color tone changes and fast motions, which show the passage of time.
But time-lapse videos usually require a sophisticated hardware setup and are time-consuming to capture and edit.
Therefore, it is desirable and helpful to design and develop a 
system to facilitate the creation of time-lapse videos.

\begin{figure}
\begin{center}
\centering

    \includegraphics[width=0.19\linewidth]{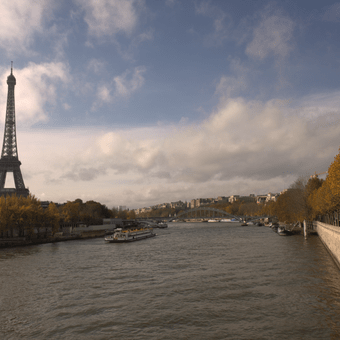}
    \includegraphics[width=0.19\linewidth]{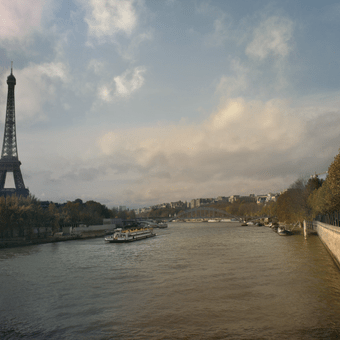}
    \includegraphics[width=0.19\linewidth]{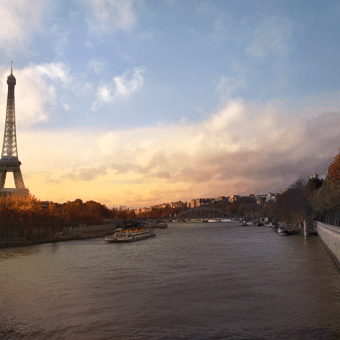}
    \includegraphics[width=0.19\linewidth]{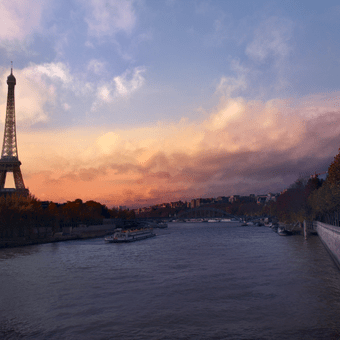}
    \includegraphics[width=0.19\linewidth]{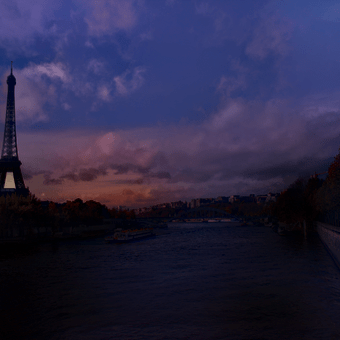}

    \includegraphics[width=0.19\linewidth]{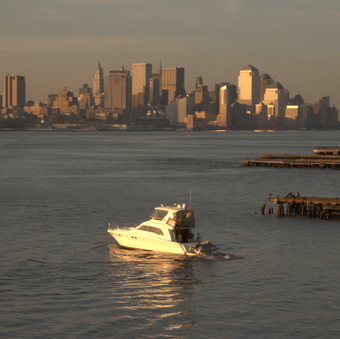}
    \includegraphics[width=0.19\linewidth]{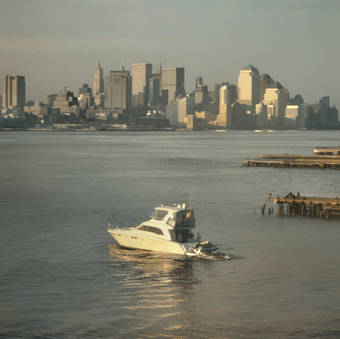}
    \includegraphics[width=0.19\linewidth]{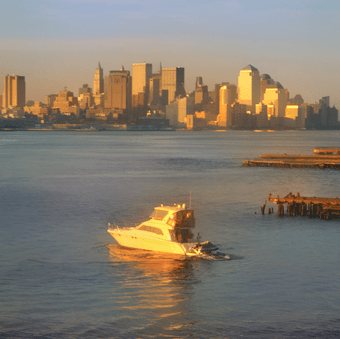}
    \includegraphics[width=0.19\linewidth]{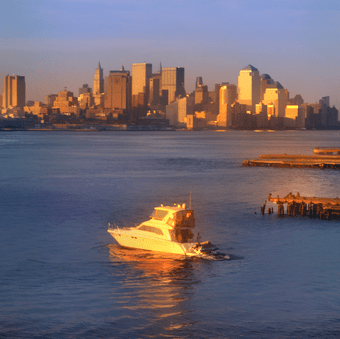}
    \includegraphics[width=0.19\linewidth]{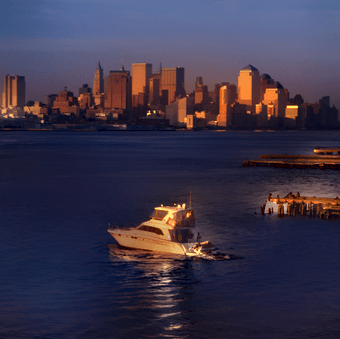}

    \includegraphics[width=0.19\linewidth]{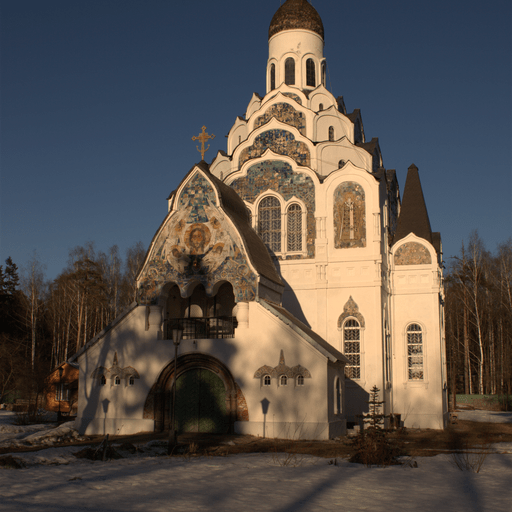}
    \includegraphics[width=0.19\linewidth]{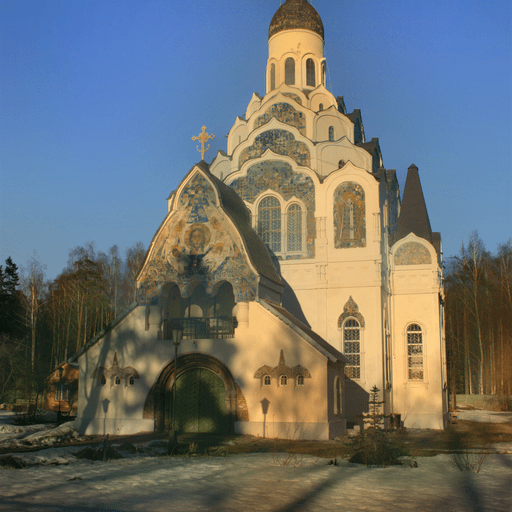}
    \includegraphics[width=0.19\linewidth]{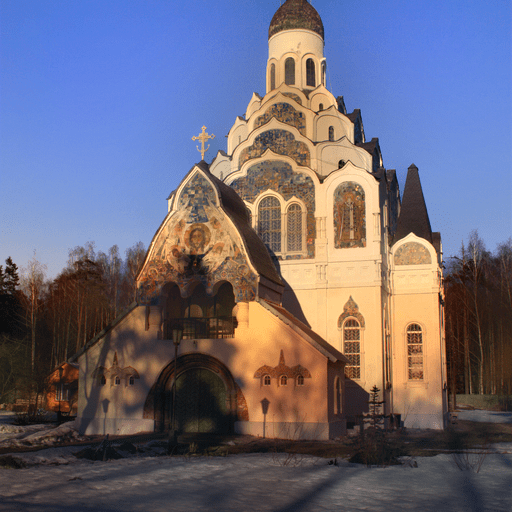}
    \includegraphics[width=0.19\linewidth]{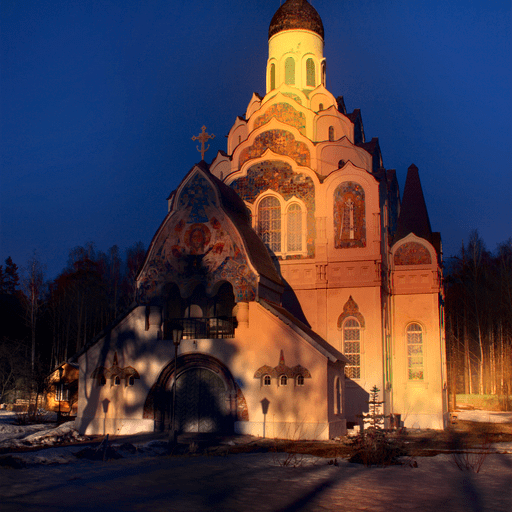}
    \includegraphics[width=0.19\linewidth]{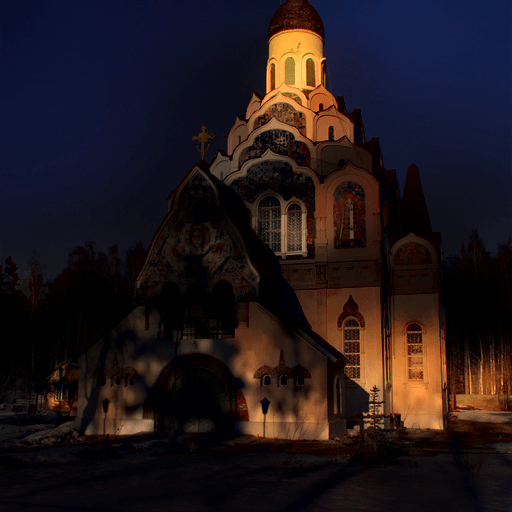}
\end{center}
\vspace{\figurevspace} 
\caption{For each single outdoor image (the first column), we can predict continuous illuminance changes over time in an end-to-end manner (four columns on the right).
}
\label{fig:teaser}
\end{figure}

The appearance of an outdoor scene depends on many complicated factors including weather, season, time of day, and objects in the scene.
As a result, most time-lapse videos present highly nonlinear changes in both the temporal and spatial domains, and it is difficult to derive an explicit model to synthesize realistic time-lapse videos while taking all the deciding factors into account accurately.
With various emerging social network services, a large amount of time-lapse video data that is captured at different locations around the world has become accessible on the Internet.
Therefore, a natural idea for generating time-lapse videos is to automatically synthesize the animation output by learning from a large-scale video database.
A data-driven hallucination algorithm~\cite{Shih:2013:DHD} was proposed to synthesize a time-lapse video from an input image via a color transfer based on a reference video retrieved from a database.
However, this framework needs to store the entire database of time-lapse videos for retrieval at runtime. Also, it may not always be possible to find a reference video that has components semantically similar to the input image for a visually plausible color transfer.
Recent advances in computer vision and machine learning have shown that deep neural networks can be used to achieve photorealistic style transfer~\cite{Luan:2017:DPS,Li:2018:CFS,Lee:2018:DIT} and to synthesize high-fidelity video sequences~\cite{Kim:2018:DVP,Wang:2018:vid2vid,Bansal:2018:Recycle-GAN}.
Yet most existing deep video generation techniques require a reference video or a label map sequence to guide the synthesis of the output video.

In this work, we present an end-to-end data-driven time-lapse hallucination solution for a single image without the requirement of any semantic labels or reference videos at runtime.
Given an outdoor image as the input, our method can automatically predict how the same scene will look like at different times of a day and generate a time-lapse video with continuous and photorealistic illumination changes by using the timestamp as the control variable.
See \Fref{fig:teaser} for some example results generated by our system.

Conventionally, video generation tasks have been modelled by spatiotemporal methods such as recurrent neural networks and volumetric convolutions~\cite{Xiong:2018:MDGAN,Vondrick:2016:GVS,Tulyakov:2018:MoCoGAN,Zhou:2016:LTR}.
However, it is challenging to achieve our goal with these approaches since the raw footage of existing time-lapse datasets~\cite{Jacobs:2007:AMOS,Shih:2013:DHD} contains a number of unwanted camera motions, moving objects, or even corrupted frames, which aggravates the quality of output sequences.
In this work, we cast our task as a conditional image-to-image translation task using the timestamp as the control variable, which enables our learning to be robust to such outliers through the structure preserving property~\cite{Zhu:2017:CycleGAN,Isola:2017:pix2pix}.
However, this alone cannot generate plausible time-lapse videos due to the independent modeling of different times.
To effectively train the continuous change of illumination over time, we propose a multi-frame joint conditional generation framework (\Sref{sec:algorithm_multi_frame}).
For training, we leverage the AMOS dataset~\cite{Jacobs:2007:AMOS} and build a large collection of outdoor images with the corresponding timestamps of when the photos were taken.


One issue of using the AMOS dataset is that many footages in the dataset are visually uninteresting, because the dataset is collected from hundreds of thousands of surveillance cameras capturing outdoor scenes such as highways and landscapes.
To further improve the visual quality of our synthesis output, we also leverage the time-lapse video database TLVDB~\cite{Shih:2013:DHD}, which is a small collection of time-lapse videos.
The videos in the TLVDB dataset present rich illumination changes but do not have the ground-truth timestamp for each frame.
To jointly learn from both the TLVDB dataset and the AMOS dataset, we propose a multi-domain training scheme (\Sref{sec:algorithm_multi_domain}) based on image domain translation~\cite{Zhu:2017:CycleGAN,Isola:2017:pix2pix}.
It enables the TLVDB dataset to be trained with our conditional generation framework in a {\em semi-supervised} manner, which removes the necessity for timestamps in the TLVDB dataset.
Our training scheme also effectively handles the difference of data distribution between the two datasets and makes the training process more stable compared to a na\"ive implementation.

We show a variety of time-lapse video synthesis results on diverse input images and compare our method with alternative approaches (\Sref{sec:results}).
We also verify the design and implementation of our framework via extensive ablation studies and evaluations\commentout{ (\Sref{sec:evaluations})}.
In summary, our contributions are:
\begin{itemize}
\item We present the first solution for synthesizing a time-lapse video with continuous and photorealistic illumination changes from a single outdoor image without the requirement of any reference video at runtime.
\item We propose a multi-frame joint conditional network to learn the distributions of color tones at different times of a day while bypassing the motions and outliers in the training data.
\item We propose a multi-domain training scheme for stable semi-supervised learning from different datasets to further improve the visual quality of synthesis output.
\end{itemize}

\section{Related Work}
\label{sec:related_work}

\paragraph{Image and video stylization.}

Image and video stylization has been an active research area over the past few years, especially with the recent advances in deep neural networks for robust and effective computation of visual features~\cite{Gatys:2016:IST,Johnson:2016:PLR,Zhu:2017:CycleGAN,Luan:2017:DPS,Li:2018:CFS,Lee:2018:DIT}.
A typical usage scenario of visual stylization algorithms is to transfer the style of the input from one \emph{source} domain into another \emph{target} domain while keeping the content, such as night to day, sketch to photo, label map to image, or vice versa~\cite{Karacan:2016:AGAN,Yi:2017:DualGAN,Isola:2017:pix2pix,Chen:2017:PIS,Wang:2018:vid2vid,Bansal:2018:Recycle-GAN,Anoosheh:2018:NIT}.
In contrast to these prior methods, our technique aims to change the illumination of an input image in a \emph{continuous} manner by using time of day as the control variable for a conditional generative model.

\paragraph{Animating still images.}

Creating animation from a single image has been a longstanding research problem in computer vision and computer graphics.
Early work on this topic relies on either user interactions~\cite{Chuang:2005:APS} or domain-specific knowledge~\cite{Xu:2008:AAM,Jhou:2016:ASL}.
Most related to our approach, a data-driven hallucination method~\cite{Shih:2013:DHD} was proposed to synthesize a time-lapse video from a single outdoor image by a color transfer algorithm based on a reference video. 
On the contrary, we only need to store a compact model for the synthesis and do not require any reference video at runtime.
Therefore our method requires much less storage and can be more robust for input images that are significantly different from all the available reference videos.

More recently, deep neural networks such as generative adversarial networks (GANs) and variational autoencoders (VAEs) have been widely used for video synthesis and future frame prediction~\cite{Xue:2016:VDP,Walker:2016:AUF,Vondrick:2016:GVS,Villegas:2017:DMC,Tulyakov:2018:MoCoGAN}.
Due to the limited capability of neural networks, most of these techniques can only generate very short or fixed-length sequences with limited resolution, and/or have been focusing on specific target phenomenon, such as object transformation~\cite{Zhou:2016:LTR} and cloud motions~\cite{Xiong:2018:MDGAN}.
Our approach is complementary to these prior methods, and we can animate a variety of high-resolution outdoor images by continuously changing the color tone to generate output videos of arbitrary length.


\paragraph{Learning from video dataset.}

Compared to traditional image datasets such as ImageNet~\cite{Deng:2009:ImageNet} and COCO~\cite{Lin:2014:COCO}, large-scale video datasets (or image sequences from static cameras) usually contain rich hidden information among the coherent frames within each sequence.
On the other hand, these datasets present additional challenges to the learning algorithms since the amount of data is usually prohibitively large and less structured.
The Archive of Many Outdoor Scenes (AMOS) dataset~\cite{Jacobs:2007:AMOS} contains millions of outdoor images captured with hundreds of webcams.
In their work, the authors demonstrate the possibility of analyzing the dataset with automatic annotations, such as semantic labels, season changes, and weather conditions.
It is also possible to extract illumination, material and geometry information from time-lapse videos as shown in previous methods~\cite{Matusik:2004:PRF,Sunkavalli:2007:FTV,Laffont:2015:IDI}.
Most recently, Li and Snavely~\cite{Li:2018:LII} proposed to learn single-view intrinsic image decomposition from time-lapse videos in the wild without ground truth data.
We draw inspirations from this line of research and propose to learn a generative model for the time-lapse video synthesis.

\section{Our Method}
\label{sec:method}



\paragraph{Problem statement.}

To synthesize a time-lapse video from a single input image, we define our task as conditional image translation based on generative adversarial networks (GANs)~\cite{Goodfellow:2014:GAN,Mirza:2014:cGAN} by using the time of day as the conditional variable.
Formally, let $\inputimage$ be an input image and $\timestamp \in \mathbb{R}$ be the target timestamp variable in the range of $[0, \, 1)$ for a whole day.
Our task can then be described as $\outputimage_{\timestamp}=\generator(\inputimage, \timestamp)$ where a generator $\generator$ hallucinates the color tone of the input $\inputimage$ to predict the output image $\outputimage_{\timestamp}$ at the time $\timestamp$\commentout{ with a random latent variable $\latentz$ sharing for all $\timestamp$.
The latent variable $\latentz$ models diverse color tone change for a day}.
To generate a time-lapse video, we sample a finite set of timestamps $\timestamp \in \{\timestamp_0, \timestamp_1, \timestamp_2, \cdots, \timestamp_n\}$, then aggregate generated images to form a video $\video =\{\outputimage_0, \outputimage_1, \outputimage_2, \cdots, \outputimage_n\}$.

Note that our goal is to model continuous and nonlinear change of color tone over the time without considering dynamic motions such as moving objects.
At test time, we synthesize output video in an end-to-end manner without the requirement of any reference video, scene classification, or semantic segmentation.
In addition, our approach enables generating any number of frames at inference time by using real-valued $\timestamp$ as the control variable.

\paragraph{Datasets.}

\begin{figure}
\begin{center}
\includegraphics[width=\linewidth]{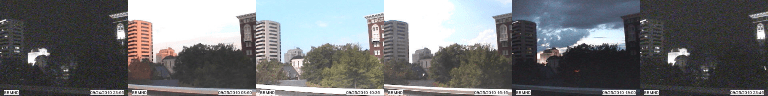}
\includegraphics[width=\linewidth]{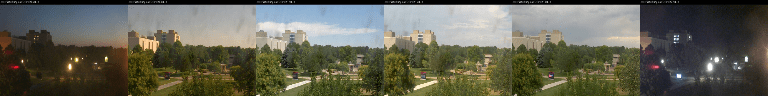}
\includegraphics[width=\linewidth]{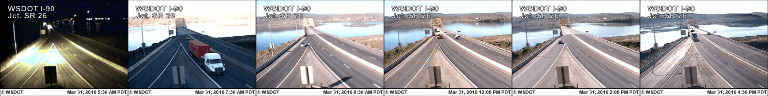}
\includegraphics[width=\linewidth]{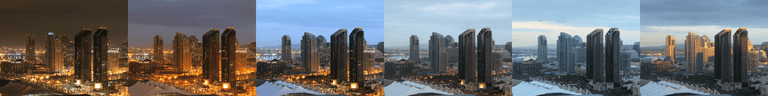}
\includegraphics[width=\linewidth]{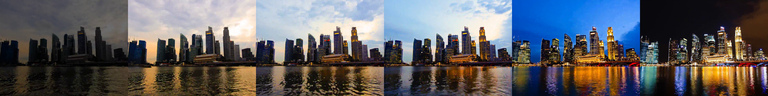}
\end{center}
\vspace{\figurevspace} 
\caption{Sample sequences from the AMOS (top three rows) and the TLVDB (bottom two rows) datasets.}
\label{fig:dataset}
\end{figure}

To learn an end-to-end model for the time-lapse video synthesis from a single input image, we leverage the AMOS dataset~\cite{Jacobs:2007:AMOS} and the time-lapse video database (TLVDB)~\cite{Shih:2013:DHD}.
The AMOS dataset is a large-scale dataset of outdoor image sequences captured from over $35,000$ webcams around the world.
A typical sequence in the AMOS dataset has tens of frames for every $24$ hours with timestamp labels of the time when they were captured.
The TLVDB dataset contains $463$ real time-lapse videos and most of them are about landmark scenes.
Each of the videos in the TLVDB dataset has at least hundreds of frames without timestamp labels and the numbers of frames are different.
\Fref{fig:dataset} shows several sample frames from the two datasets.

However, it is not easy to directly train a generative model using these two datasets since they contain many outliers, or even corrupted data.
For instance, the images in some sequences are not aligned due to camera movement and contain abrupt scene changes, text overlays, fade in/out effects, etc.
We only prune some obviously corrupted frames and sequences manually, as removing all the noisy data requires extensive human labor with heuristics.


\subsection{Multi-Frame Joint Conditional Generation}
\label{sec:algorithm_multi_frame}

We denote the AMOS dataset as $\amos$.
Each data $(\image_i, \timestamp_i)$ in $\amos$ is a pair of an image $\image_i$ and its corresponding timestamp $\timestamp_i$. 
As a na\"ive approach, it is possible to adopt a conditional \emph{image-to-image} translation framework using the timestamp $\timestamp$ as the conditional variable.
Specifically, from the AMOS dataset $\amos$, we may train a generator $\generator$ to synthesize an image $\outputimage_{\timestamp}=\generator(\inputimage, \timestamp)$ for a target timestamp $\timestamp$, while a discriminator $\discriminator$ is trained to distinguish whether a pair of an image and a timestamp is real or fake.

However, we found that training such a na\"ive model using each frame in $\amos$ independently tends to generate implausible color tone in the output sequence.
This is because
the illumination at certain time $\timestamp$ is correlated with the illumination at different times of the same day.
In addition, the illumination changes over multiple days can be different due to additional factors such as locations, season, and weather.
To this end, we propose to train multiple frames from each sequence with a shared latent variable $\latentz$.
We use this latent variable $\latentz$ as the temporal context of each sequence to learn the joint distribution of color tones at different times of the day.

\begin{figure}
\centering
\includegraphics[width=\linewidth]{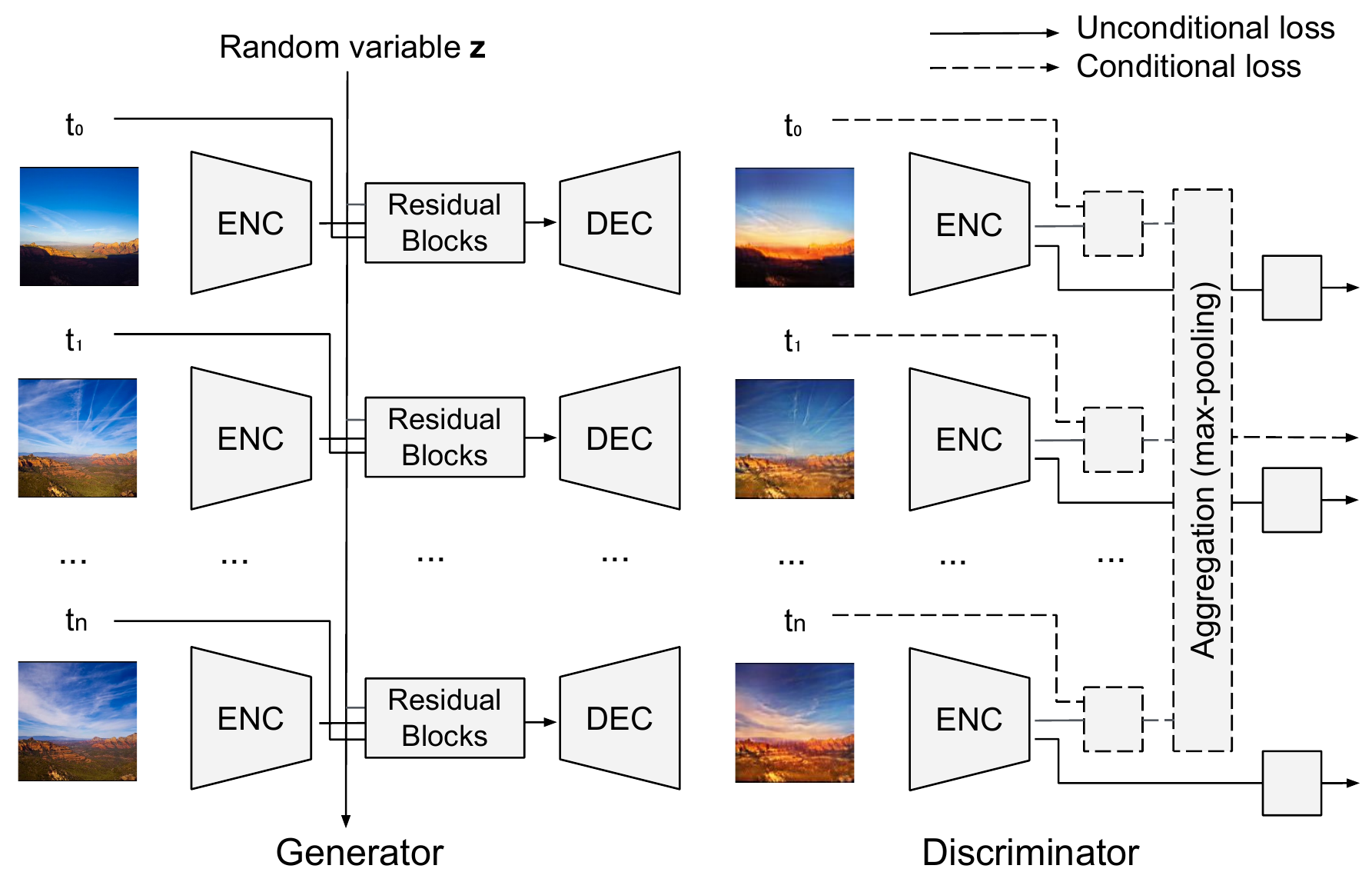}
\vspace{\figurevspace}
\caption{Illustration of our multi-frame joint conditional GAN method. For our discriminator, the encoded images are concatenated with the timestamps (dashed rectangles) before being aggregated to compute the conditional loss, while each image is directly used as an input to compute the unconditional loss (solid rectangles).
}
\label{fig:multi_frame_cgan}
\end{figure}

\paragraph{Generator.}

Our generator $\generator$ leverages a typical encoder-decoder architecture based on our proposed multi-frame joint generation scheme, as depicted in \Fref{fig:multi_frame_cgan}(left).
Let $\amosframeset$ be a set of frames sampled from the same sequence in the AMOS dataset $\amos$:
\begin{equation}
\amosframeset = \{(\image_0, \timestamp_0), (\image_1, \timestamp_1), (\image_2, \timestamp_2), \cdots, (\image_n, \timestamp_n)\}.
\label{eq:frameset_amos_def}
\end{equation}
An input image $\inputimage_i$ is encoded by the encoder $\generatorencoder$.
The shared latent variable $\latentz$ is sampled from the standard normal distribution $\normaldistribution(0, 1)$ to represent the temporal context of the sequence from which $\amosframeset$ is sampled.
Then several residual blocks~\cite{He:2016:DRL} take the encoded image $\generatorencoder(\inputimage_i)$ together with the latent variable $\latentz$ and the timestamp $\timestamp_i$ as the input and generate output features with a new color tone.
Finally, the decoder $\generatordecoder$ in $\generator$ decodes the feature from the residual blocks into an image $\outputimage_{\timestamp_i}$ as the reconstructed output of $\image_i$:
\begin{equation}
\begin{split}
\outputimage_{\timestamp_i} & = \generator(\inputimage_i, \timestamp_i, \latentz)\\
& = \generatordecoder(\generatorencoder(\inputimage_i), \timestamp_i, \latentz),
\end{split}
\label{eq:image_reconstruction_output}
\end{equation}
where we omit the residual blocks for simplicity.
The entire reconstructed output of $\amosframeset$ consists of all the generated frames:
\begin{equation}
\begin{split}
\fakeamosframeset = \{\outputimage_{\timestamp_0}, \outputimage_{\timestamp_1}, \outputimage_{\timestamp_2}, \cdots, \outputimage_{\timestamp_n}\}.
\end{split}
\label{eq:frameset_amos_output_def}
\end{equation}

During the training, we use different input images from the same sequence as depicted in~\Fref{fig:multi_frame_cgan}, which enables $\generator$ to ignore moving factors.
At the inference time, we use the same input image multiple times to get an output sequence.

\paragraph{Discriminator.}

Our discriminator $\discriminator$ is divided into two parts, i.e., an \emph{unconditional} discriminator $\unconditionaldiscriminator$ for each individual output image $\outputimage$ and a \emph{conditional} discriminator $\conditionaldiscriminator$ for the set of the reconstructed images from an input frame set $\amosframeset$.
The unconditional discriminator $\unconditionaldiscriminator$ is used to differentiate if each individual image is real or fake.
The conditional discriminator $\conditionaldiscriminator$ distinguishes if a generated frame set $\fakeamosframeset$ is a real time-lapse sequence.
In other words, $\conditionaldiscriminator$ checks not only whether each individual frame $\outputimage_{\timestamp_i}$ matches the corresponding $\timestamp_i$, but also whether $\fakeamosframeset$ presents realistic color tone changes over time.

We train both $\unconditionaldiscriminator$ and $\conditionaldiscriminator$ based on the same image encoder $\discriminatorencoder$, as shown in \Fref{fig:multi_frame_cgan}(right).
For the conditional discriminator $\conditionaldiscriminator$, the encoded image and the corresponding timestamp are concatenated for each frame and all the frames from the same sequence are aggregated to compute the discriminator score.
Since the input of the conditional discriminator $\conditionaldiscriminator$ is an unordered set of $\{(\outputimage_{\timestamp_i}, \timestamp_i)\}$ rather than an ordered sequence, the discriminator score should be permutation-invariant~\cite{Zaheer:2017:DeepSets}.
Therefore, we use {\em max-pooling} to aggregate encoded features of multiple frames.

\paragraph{Adversarial losses.}
The adversarial losses of our multi-frame joint conditional generation algorithm consist of an unconditional loss and a conditional loss.
The unconditional adversarial loss $\unconditionalloss$ can be formally described as:
\begin{equation}
\begin{split}
\unconditionalloss &= \objectivefunc_{\inputimage \sim \amos}[\log{\unconditionaldiscriminator(\inputimage)}] \\
&+ \objectivefunc_{(\inputimage, \timestamp) \sim \amos, \latentz \sim \normaldistribution}[1 - \log{\unconditionaldiscriminator(\generator(\inputimage, \timestamp, \latentz))}],
\end{split}
\label{eq:multi_frame_loss_uncond}
\end{equation}
where $\normaldistribution$ is the standard normal distribution.
Our conditional adversarial loss $\conditionalloss$ is defined as below:
\begin{equation}
\begin{split}
\conditionalloss &= \objectivefunc_{\frameset_\amos \sim \amos}[\log{\conditionaldiscriminator(\frameset_\amos)}] \\
&+ \objectivefunc_{\fakeamospairset \sim \amos}[1 - \log{\conditionaldiscriminator(\fakeamospairset)}] \\
&+ \objectivefunc_{\frameset_\amos \sim \amos, \latentz \sim \normaldistribution}[1 - \log{\conditionaldiscriminator(\generator(\frameset_\amos, \latentz))}].
\end{split}
\label{eq:multi_frame_loss_cond}
\end{equation}
To effectively train the correlation between the input image $\inputimage_i$ and its corresponding timestamp $\timestamp_i$, we introduce an additional term as shown in the second row of~\eref{eq:multi_frame_loss_cond} for a set of negative pairs $\fakeamospairset$, which we collect by randomly sampling mismatched pairs from $\amosframeset$:
\begin{equation}
\fakeamospairset = \{(\image_i, \timestamp_j) \mid i \neq j\}.
\label{eq:fakepairset_amos_def}
\end{equation}

\subsection{Multi-Domain Training}
\label{sec:algorithm_multi_domain}

Our multi-frame joint conditional generation method effectively captures diverse illumination variation over time.
However, the model trained based on the AMOS dataset alone tends to generate uninteresting outputs such as clipped and less saturated colors especially in the sky region, since most footages of the AMOS dataset were captured by surveillance cameras.
To further improve the visual quality of synthesis output, we propose to additionally leverage the TLVDB dataset~\cite{Shih:2013:DHD} and denote it as $\tlvdb$.
Most videos in $\tlvdb$ are about landmark scenes captured using professional cameras and thus present much more interesting color tone distributions and changes over time.

However, the footages in the TLVDB dataset are videos without any ground-truth timestamp label for each frame.
Therefore, it is infeasible to directly learn from this dataset using our conditional generation method described in \Sref{sec:algorithm_multi_frame}.
Furthermore, we have found that simply merging the AMOS dataset and the TVLDB dataset to train the unconditional image discriminator in~\eref{eq:multi_frame_loss_uncond} does not improve the results, due to the domain discrepancy of the two datasets.
To handle the issues of missing timestamps and inconsistent data distributions, we propose a multi-domain training method.

\begin{figure}
\centering
\includegraphics[width=\linewidth]{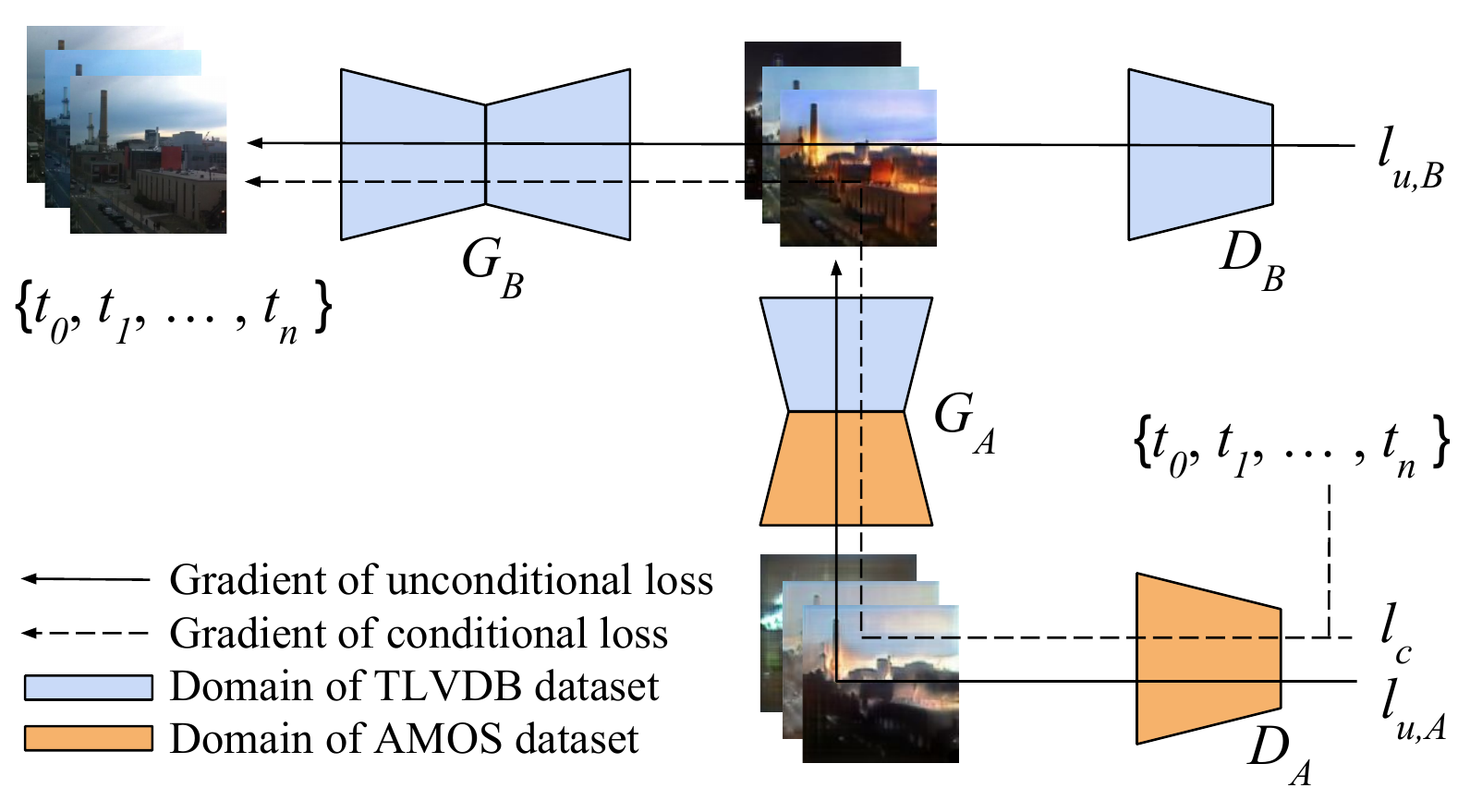}
\vspace{\figurevspace}
\caption{Illustration of our multi-domain training scheme.
}
\label{fig:multi_domain_training}
\end{figure}

Our key idea is to synthesize time-lapse sequences using the TLVDB dataset $\tlvdb$ and to learn continuous illumination changes over time from the AMOS dataset $\amos$.
\Fref{fig:multi_domain_training} shows the overview of our multi-domain training algorithm.
Basically, we train a generator $\generatortlvdb$ together with a discriminator $\discriminatortlvdb$ based on $\tlvdb$ to synthesize time-lapse sequences.
The synthesis results are then translated into the domain of $\amos$ by using another generator $\generatoramos$ as a proxy to get conditional training signals from the discriminator $\discriminatoramos$ trained on $\amos$.
With the training signal from $\discriminatoramos$, $\generatortlvdb$ can be trained to synthesize images with the timestamp $\timestamp$ being taken into account.
To this end, we adopt our multi-frame conditional generation network (\Sref{sec:algorithm_multi_frame} and \Fref{fig:multi_frame_cgan}) for $\generatortlvdb$ and $\discriminatoramos$, while using vanilla DCGAN~\cite{Radford:2015:DCGAN} and U-Net~\cite{Ronneberger:2015:UNet} for $\discriminatortlvdb$ and $\generatoramos$, respectively.

\paragraph{Loss functions.}
For our multi-domain training scheme, the unconditional loss in \eref{eq:multi_frame_loss_uncond} is reformulated as:
\begin{equation}
\begin{split}
\unconditionalloss &= \unconditionallossamos + \unconditionallosstlvdb,
\end{split}
\end{equation}
where
\begin{equation}
\begin{split}
\unconditionallossamos &= \objectivefunc_{\inputimage \sim \amos}[\log{\discriminatoramos(\inputimage)}] \\
&+ \objectivefunc_{\inputimage \sim \tlvdb, \timestamp \sim \amos, \latentz \sim \mathcal{N}}[1 - \log{\discriminatoramos(\generator_\amos(\generator_\tlvdb(\inputimage, \timestamp, \latentz)))}],
\end{split}
\end{equation}
and
\begin{equation}
\begin{split}
\unconditionallosstlvdb &= \objectivefunc_{\inputimage \sim \tlvdb}[\log{\discriminatortlvdb(\inputimage)}] \\
&+ \objectivefunc_{\inputimage \sim \tlvdb, \timestamp \sim \amos, \latentz \sim \normaldistribution}[1 - \log{\discriminatortlvdb(\generator_\tlvdb(\inputimage, \timestamp, \latentz))}].
\end{split}
\end{equation}
The conditional loss in \eref{eq:multi_frame_loss_cond} can be rewritten as:
\begin{equation}
\begin{split}
\conditionalloss &= \objectivefunc_{\frameset_\amos \sim \amos}[\log{\discriminatoramos(\frameset_\amos)}] \\
&+ \objectivefunc_{\fakeamospairset \sim \amos}[1 - \log{\discriminatoramos(\fakeamospairset)}] \\
&+ \objectivefunc_{\tlvdbframeset \sim \tlvdb, \latentz \sim \normaldistribution}[1 - \log{\discriminatoramos(\generatoramos(\generatortlvdb(\tlvdbframeset, \latentz)))}],
\end{split}
\end{equation}
where we omit the exact definition of $\frameset_\tlvdb$ for simplicity.
We also add a reconstruction loss $\reconstructionloss$ for $\generatoramos$ based on $L1$ norm to enforce the network to learn the mapping from a sample in one domain to a similar one in another domain:
\begin{equation}
\reconstructionloss = \Big\Vert \generatoramos(\generatortlvdb(\inputimage, \timestamp, \latentz)) - \generatortlvdb(\inputimage, \timestamp, \latentz) \Big\Vert_1.
\label{eq:reconstruction_loss_def}
\end{equation}


\begin{algorithm}[t]
\caption{Our training algorithm}
\begin{algorithmic}
\STATE Set the learning rate $\learningrate$
\STATE Initialize network parameters $\networkparams_{\generatoramos}$, $\networkparams_{\discriminatoramos}$, $\networkparams_{\generatortlvdb}$, $\networkparams_{\discriminatortlvdb}$
\FOR{the number of iterations}
\STATE Sample $\amosframeset, \fakeamospairset \sim \amos$, $\tlvdbframeset \sim \tlvdb$
\STATE \textbf{(1) Update the discriminators $\discriminatoramos$ and $\discriminatortlvdb$}
\STATE Sample $\latentz \sim \normaldistribution$
\STATE Generate $\generatortlvdb(\tlvdbframeset, \latentz)$, $\generatoramos(\generatortlvdb(\tlvdbframeset, \latentz))$
\STATE $\networkparams_{\discriminatoramos} = \networkparams_{\discriminatoramos} + \learningrate \nabla_{\networkparams_{\discriminatoramos}} (\unconditionallossamos + \conditionalloss)$
\STATE $\networkparams_{\discriminatortlvdb} = \networkparams_{\discriminatortlvdb} + \learningrate \nabla_{\networkparams_{\discriminatortlvdb}} \unconditionallosstlvdb$
\STATE \textbf{(2) Update the generator $\generatoramos$}
\STATE Sample $\latentz \sim \normaldistribution$
\STATE Generate $\generatortlvdb(\tlvdbframeset, \latentz)$, $\generatoramos(\generatortlvdb(\frameset_\tlvdb, \latentz))$
\STATE $\networkparams_{\generatoramos} = \networkparams_{\generatoramos} + \learningrate \nabla_{\networkparams_{\generatoramos}} (\unconditionallossamos + \reconstructionloss)$
\STATE \textbf{(3) Update the generator $\generatortlvdb$}
\STATE Sample $\latentz \sim \normaldistribution$
\STATE Generate $\generatortlvdb(\tlvdbframeset, \latentz)$, $\generatoramos(\generatortlvdb(\frameset_\tlvdb, \latentz))$
\STATE $\networkparams_{\generatortlvdb} = \networkparams_{\generatortlvdb} + \learningrate \nabla_{\networkparams_{\generatortlvdb}} (\unconditionallosstlvdb + \conditionalloss)$
\ENDFOR
\end{algorithmic}
\label{alg:training}
\end{algorithm}

\paragraph{Training algorithm.}
Our network is trained by solving the following minimax optimization problem:
\begin{equation}
\generatoramos^{*}, \discriminatoramos^{*}, \generatortlvdb^{*}, \discriminatortlvdb^{*} = \min_{\generatoramos, \generatortlvdb} \max_{\discriminatoramos, \discriminatortlvdb} \unconditionalloss + \conditionalloss + \lambda \, \reconstructionloss,
\end{equation}
where $\lambda$ is the weight of the reconstruction loss $\reconstructionloss$ defined in~\eref{eq:reconstruction_loss_def}.
Note that $\conditionalloss$ is only used for updating $\generatortlvdb$ and we do not update $\generatoramos$ with the gradient from $\conditionalloss$, since the purpose of using $\generatoramos$ is to translate the domain without considering the timestamp condition.
In addition, we update $\generatoramos$ and $\generatortlvdb$ alternately as they are dependent on each other.
Our training procedure is described in \Aref{alg:training} step by step.


\subsection{Guided Upsampling}
\label{sec:algorithm_upsampling}

Since our training data is very limited and contains lots of noise, it is difficult to train the network to directly output full-resolution results while completely preserving the local structure in the input image.
Therefore, we first train our generative network and predict output $\outputimage$ at a lower resolution.
Then we apply an automatic guided upsampling approach, following the local color transfer method in~\cite{He:2017:NCT}, as a post-processing step to obtain the full-resolution result.

\begin{figure}[t]
  \centering

    \includegraphics[width=0.19\linewidth]{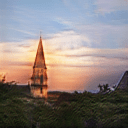}
    \includegraphics[width=0.38\linewidth]{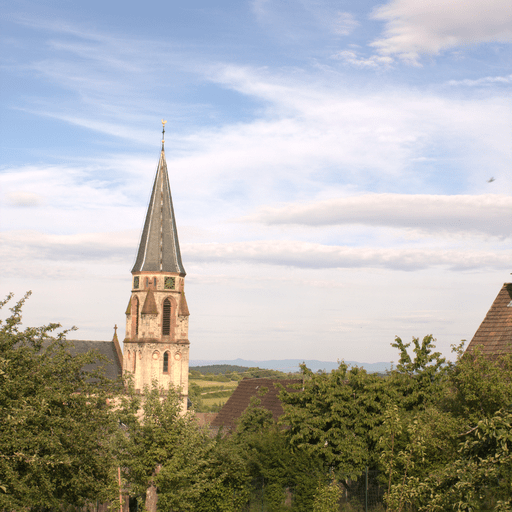}
    \includegraphics[width=0.38\linewidth]{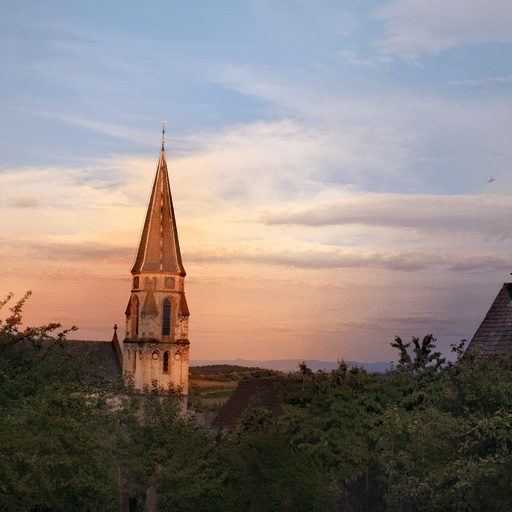}

    \includegraphics[width=0.19\linewidth]{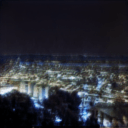}
    \includegraphics[width=0.38\linewidth]{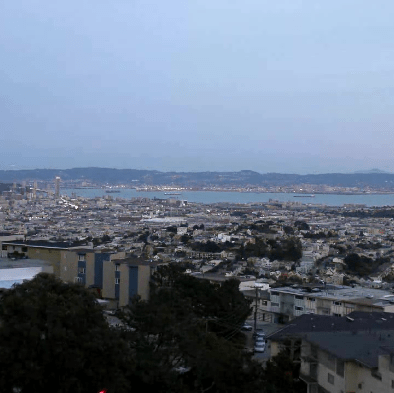}
    \includegraphics[width=0.38\linewidth]{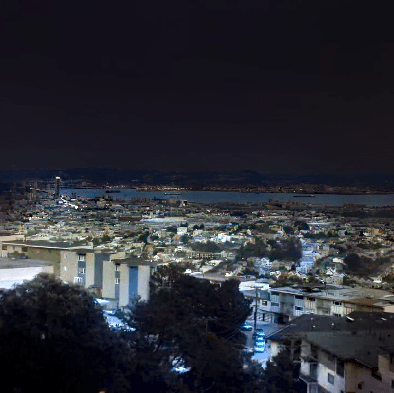}
\caption{We apply guided upsampling in a post-processing step to get full-resolution output. From left to right: the output predicted by our network, the original input, and the upsampling result.}
\label{fig:guided_upsampling}
\end{figure}

Basically, we model the per-pixel linear transformation between the final result $\finalimage$ and the input image $\inputimage$ at a pixel location $\pixel$ as a scaling factor $\scalingfactor(\pixel)$ with a bias $\bias(\pixel)$:
\begin{equation}
\finalimage(\pixel) = \lineartransform(\pixel)(\inputimage(\pixel)) = \scalingfactor(\pixel) \times \inputimage(\pixel) + \bias(\pixel).
\label{eq:linear_color_model}
\end{equation}
The key idea of guided upsampling is to use the raw network output $\outputimage$ as the guidance to compute the transformation $\lineartransform=\{\scalingfactor, \bias\}$, while using color information between neighboring pixels in the input image $\inputimage$ as regularization.
Specifically, we formulate the task as the following least-squares problem:
\begin{equation}
\begin{split}
\energyfunc &= \energyfunc_d + \smoothnessweight \, \energyfunc_s,\\
\energyfunc_d &= \sum_{\pixel}\big\|\finalimage(\pixel)-\outputimage(\pixel)\big\|^2,\\
\energyfunc_s &= \sum_{\pixel}\sum_{\anotherpixel \in \neighborhood(\pixel)} \pixelweight\big(\inputimage(\pixel), \inputimage(\anotherpixel)\big) \big\|\lineartransform(\pixel)-\lineartransform(\anotherpixel)\big\|^2,
\end{split}
\label{eq:upsampling_energy}
\end{equation}
where $\neighborhood(\pixel)$ is the one-ring neighborhood of $\pixel$ and $\pixelweight\big(\inputimage(\pixel), \inputimage(\anotherpixel)\big)$ measures the inverse color distance between two neighboring pixels $\pixel$ and $\anotherpixel$ in the original image $\inputimage$.
The data term $\energyfunc_d$ in \eref{eq:upsampling_energy} preserves the color from $\outputimage$ and the smoothness term $\energyfunc_s$ enforces local smoothness of the linear transformation $\lineartransform$ between neighboring pixels of similar colors.
A global constant weight $\smoothnessweight$ is used to balance the two energy terms.
We compute the least-squares optimization for each color channel independently and then upsample $\lineartransform$ bilinearly to the full resolution before apply to the original image.
See \Fref{fig:guided_upsampling} for two example results.



\section{Experiments}
\label{sec:results}

\subsection{Experimental Setup}

\paragraph{Dataset.}
We use both the AMOS~\cite{Jacobs:2007:AMOS} and the TLVDB~\cite{Shih:2013:DHD} datasets to train our network.
For the AMOS dataset, we only select sequences with geolocation information and adjust all timestamp labels to local time accordingly.
In addition, we remove some obviously corrupted data such as zero-byte images, grayscale images, etc.
All in all, we collected $40,537$ sequences containing $1,310,375$ images from $1,633$ cameras.
We split the collected AMOS dataset into a training set and a test set, which contain sequences from $1,533$ and $100$ cameras, respectively.
For the TLVDB dataset, we use $463$ videos that have $1,065,427$ images without preprocessing.
We randomly select $30$ videos as the test set and use the remaining videos for training.

\paragraph{Implementation details.}
We implement our method using PyTorch.
We train our model with $60,000$ iterations using Adam optimizer with the momentum set to $0.5$.
The batch size is set to be $4$ and the learning rate is $0.0002$.
We use $16$ frames for each example in a batch to train our multi-frame joint conditional GAN and set $\lambda$ to be $0.5$ based on visual quality.
For data augmentation, we first resize images to $136 \times 136$ and then apply random affine transformation including rotation, scale, and shear followed by random horizontal flipping.
Finally, the images are randomly cropped to patches of resolution $128 \times 128$.
For the encoder of $\generatortlvdb$, we adopt a pre-trained VGG-16 network~\cite{Simonyan:2014:VGG} while all other components are trained from scratch.\footnote{Please refer to the supplementary materials for more details.}

\paragraph{Baselines.}
We compare our method with two existing color transfer methods of Li~\etal~\cite{Li:2018:CFS} and Shih~\etal~\cite{Shih:2013:DHD} using the source code with the default parameters provided by the authors.
As both methods need reference videos to guide the output synthesis, we additionally implement the reference video retrieval method proposed in~\cite{Shih:2013:DHD}, which is to compare the global feature of an input image with reference video frames to find the most similar video.
We use the output of a fully-connected layer in a ResNet-50 model~\cite{He:2016:DRL} as the global feature.
The model is pre-trained for a scene classification task~\cite{Zhou:2017:Places}.
We additionally use a pre-trained scene parsing network~\cite{Zhou:2017:ADE20K} to produce semantic segmentation masks used in Li~\etal's method~\cite{Li:2018:CFS}.

\newcommand\galleryfigurewidth{0.19}
\begin{figure}[t]
\begin{flushright}
 \includegraphics[width=\galleryfigurewidth\linewidth]{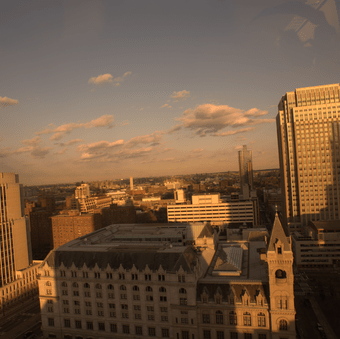}
 \includegraphics[width=\galleryfigurewidth\linewidth]{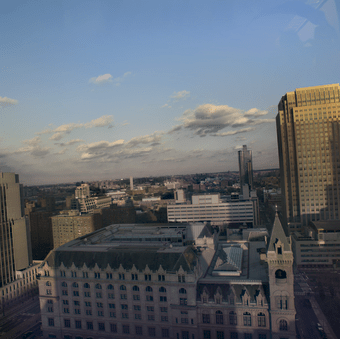}
 \includegraphics[width=\galleryfigurewidth\linewidth]{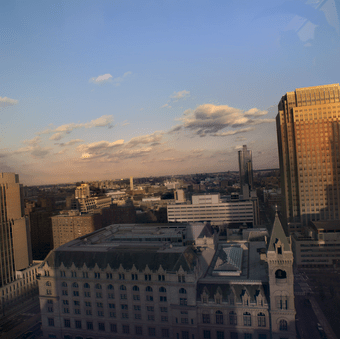}
 \includegraphics[width=\galleryfigurewidth\linewidth]{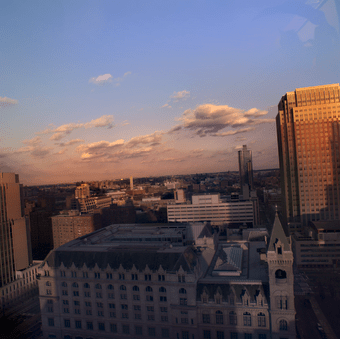}
 \includegraphics[width=\galleryfigurewidth\linewidth]{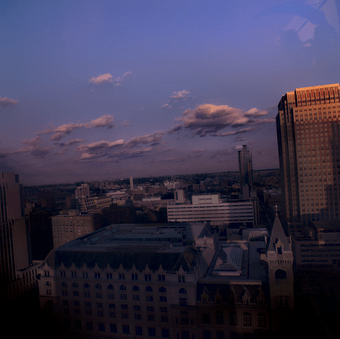}


 \includegraphics[width=\galleryfigurewidth\linewidth]{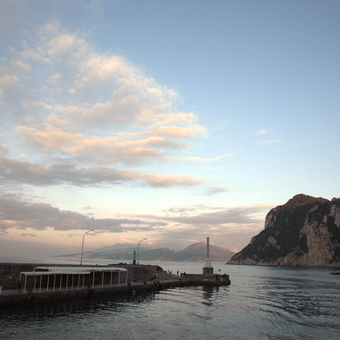}
 \includegraphics[width=\galleryfigurewidth\linewidth]{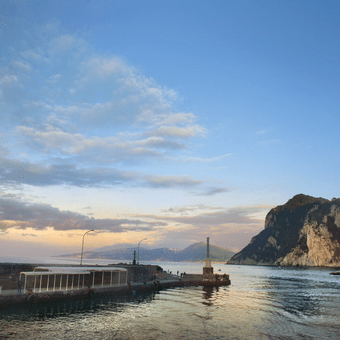}
 \includegraphics[width=\galleryfigurewidth\linewidth]{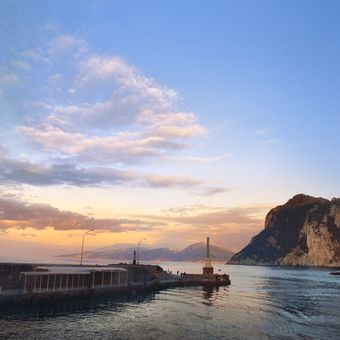}
 \includegraphics[width=\galleryfigurewidth\linewidth]{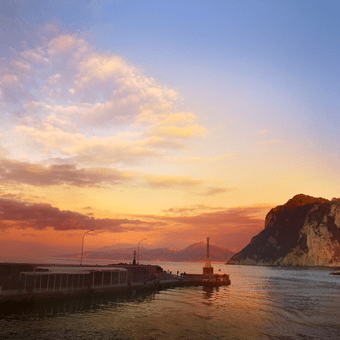}
 \includegraphics[width=\galleryfigurewidth\linewidth]{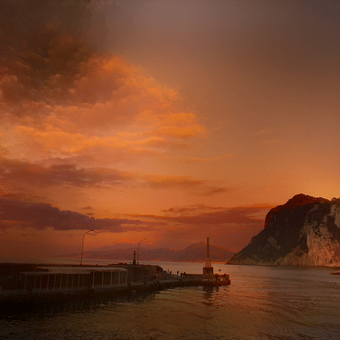}

 \includegraphics[width=\galleryfigurewidth\linewidth]{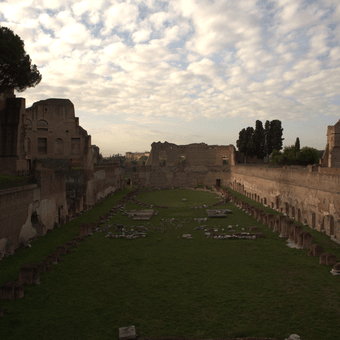}
 \includegraphics[width=\galleryfigurewidth\linewidth]{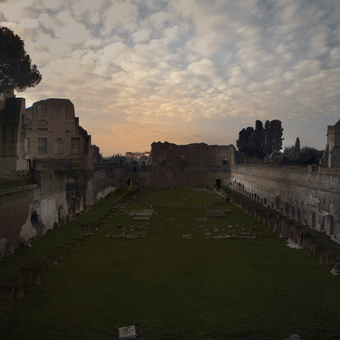}
 \includegraphics[width=\galleryfigurewidth\linewidth]{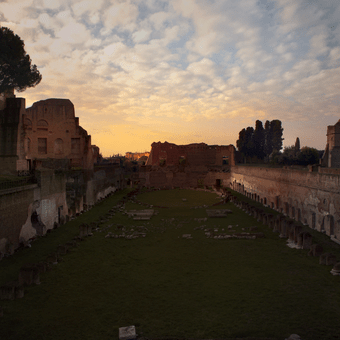}
 \includegraphics[width=\galleryfigurewidth\linewidth]{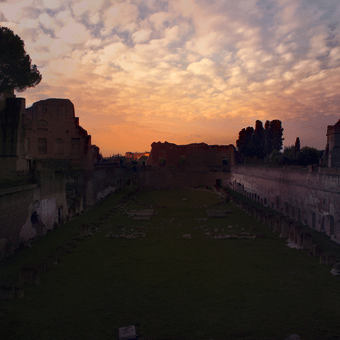}
 \includegraphics[width=\galleryfigurewidth\linewidth]{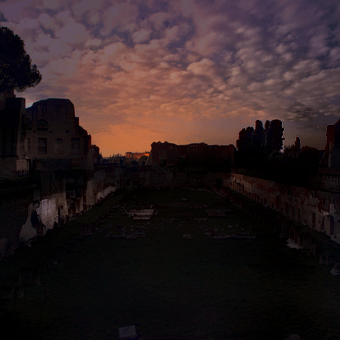}
 \small{Day}~$\xlongrightarrow{\hspace{0.55\linewidth}}$~\small{Sunset}
 \vspace{0.3\baselineskip}


 \includegraphics[width=\galleryfigurewidth\linewidth]{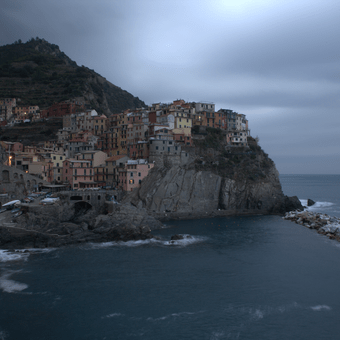}
 \includegraphics[width=\galleryfigurewidth\linewidth]{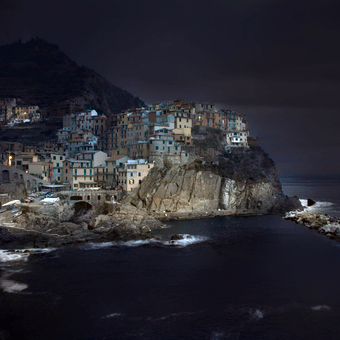}
 \includegraphics[width=\galleryfigurewidth\linewidth]{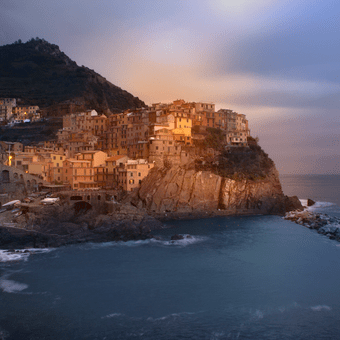}
 \includegraphics[width=\galleryfigurewidth\linewidth]{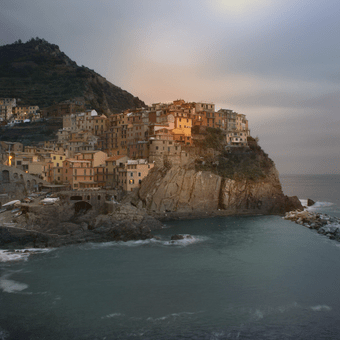}
 \includegraphics[width=\galleryfigurewidth\linewidth]{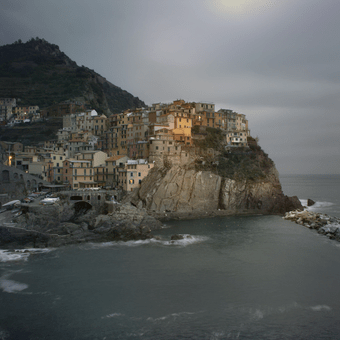}

 \includegraphics[width=\galleryfigurewidth\linewidth]{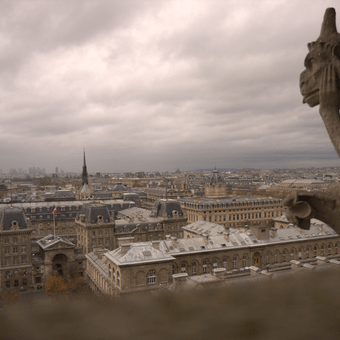}
 \includegraphics[width=\galleryfigurewidth\linewidth]{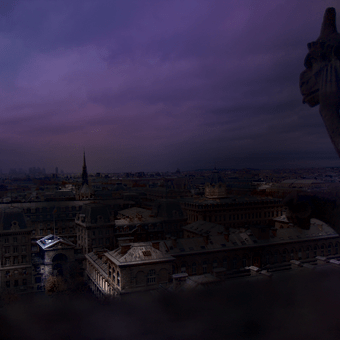}
 \includegraphics[width=\galleryfigurewidth\linewidth]{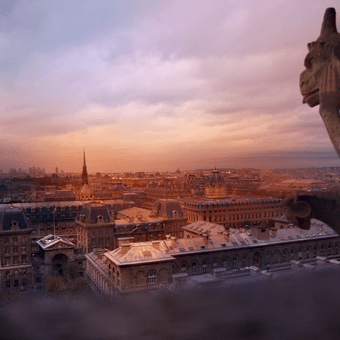}
 \includegraphics[width=\galleryfigurewidth\linewidth]{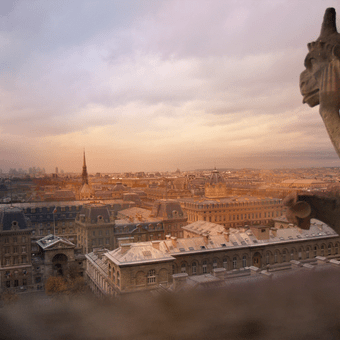}
 \includegraphics[width=\galleryfigurewidth\linewidth]{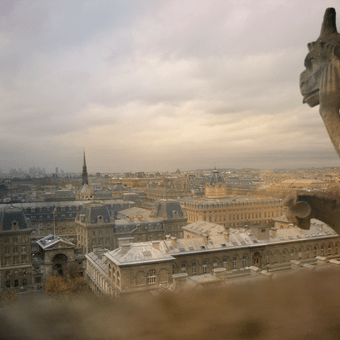}

 \includegraphics[width=\galleryfigurewidth\linewidth]{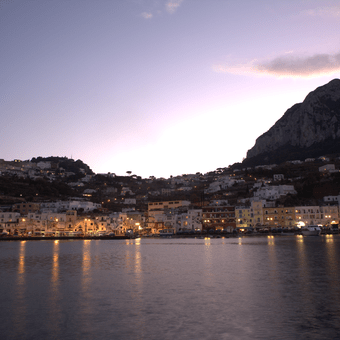}
 \includegraphics[width=\galleryfigurewidth\linewidth]{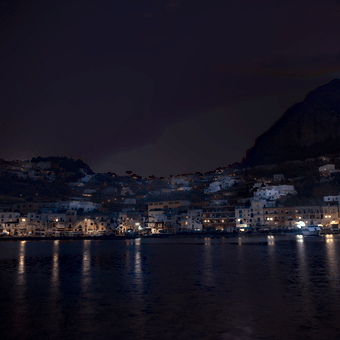}
 \includegraphics[width=\galleryfigurewidth\linewidth]{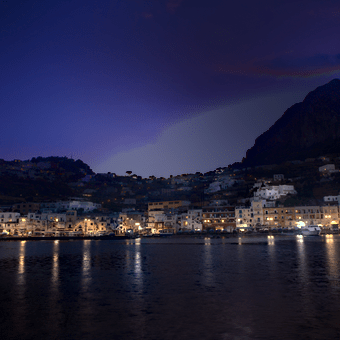}
 \includegraphics[width=\galleryfigurewidth\linewidth]{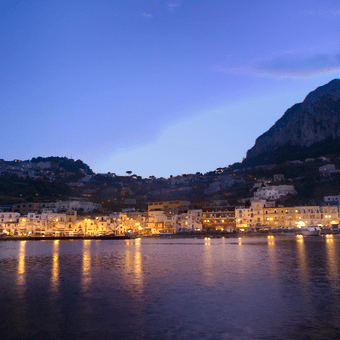}
 \includegraphics[width=\galleryfigurewidth\linewidth]{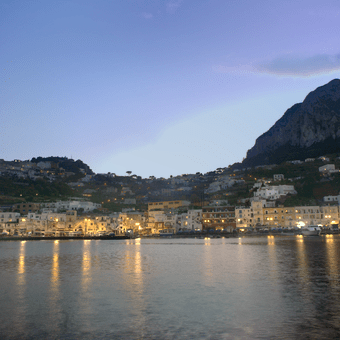}

 \small{Night}~$\xlongrightarrow{\hspace{0.55\linewidth}}$~\small{Day}
\end{flushright}
\vspace{-15pt}
\caption{Our prediction results of the input images at different times of the day. The input images are shown on the left in each row.
\snam{Again, let's choose the best results for night-to-day.}
\chongyang{(11/14/2018) Looks great to me! I picked three for the night-to-day transition and changed the text label from day-to-night to day-to-sunset. Please feel free to reverse my change.}
\menglei{I can see some artifacts around the sun overexposed region boundary, it would be great if we can find another better example, if possible.}
}
\label{fig:result_gallery}
\end{figure}

\subsection{Experimental Results}

\paragraph{Quantitative results.}
Since both baseline methods require a reference video as input while ours does not, it is difficult to conduct a completely fair comparison side-by-side.
To evaluate our method quantitatively, we performed a human evaluation following the experiment in~\cite{Shih:2013:DHD}.

Specifically, we select $24$ images from $30$ test images in the TLVDB test set and generated $24$ time-lapse sequences using the two baseline approaches~\cite{Li:2018:CFS,Shih:2013:DHD} and our method.
Then, we randomly select two or three images from each output sequence.
Eventually, we collect $71$ images for each method.
We additionally selected the same number of frames from the original videos of the test images to consider real images as another baseline.
We conducted a user study on Amazon Mechanical Turk by asking $10$ users if each image was real or fake.
We restricted the users to those who had high approval rates greater than $98\%$ to control quality.

As the result, $60.6\%$ of our results were perceived as real images by the users.
In contrast, the corresponding numbers for real images, Li \etal's method~\cite{Li:2018:CFS}, and Shih \etal's method~\cite{Shih:2013:DHD} are $67.5\%$, $34.1\%$, and $44.9\%$, respectively.
Our percentage is also higher than the reported value in~\cite{Shih:2013:DHD}, which is $55.2\%$.
We attribute the lower performance of two baseline methods to the failure of color transfer when the retrieved reference video does not perfectly match the input image.
In contrast, our results were mostly preferred by the users without using any reference video.


\newcommand\comparisonfigurewidth{0.19}
\begin{figure}[t]
\begin{center}
 \hspace{\comparisonfigurewidth\linewidth}
 \includegraphics[width=\comparisonfigurewidth\linewidth]{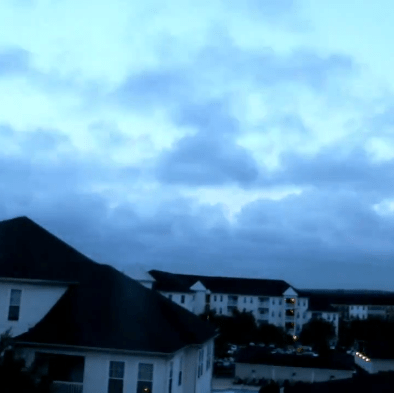}
 \includegraphics[width=\comparisonfigurewidth\linewidth]{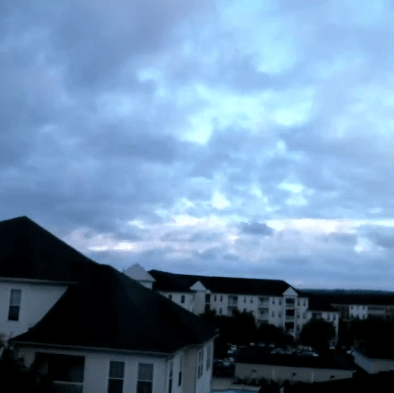}
 \includegraphics[width=\comparisonfigurewidth\linewidth]{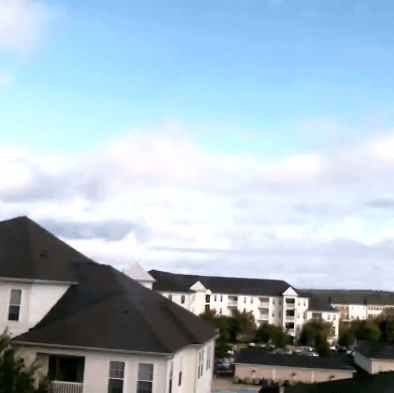}
 \includegraphics[width=\comparisonfigurewidth\linewidth]{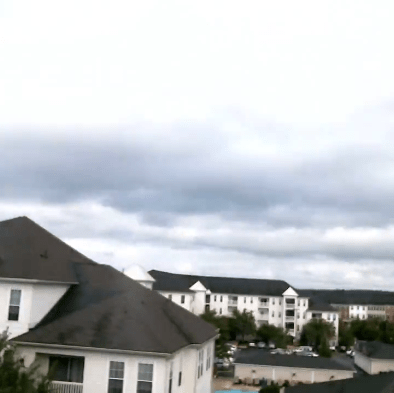}

 \includegraphics[width=\comparisonfigurewidth\linewidth]{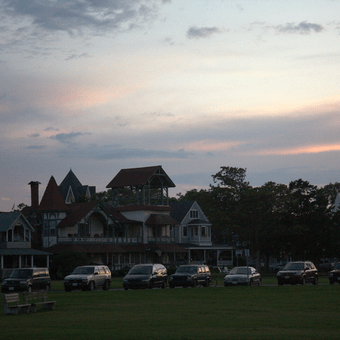}
 \includegraphics[width=\comparisonfigurewidth\linewidth]{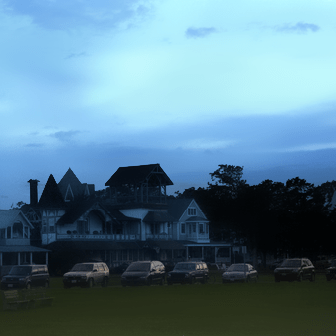}
 \includegraphics[width=\comparisonfigurewidth\linewidth]{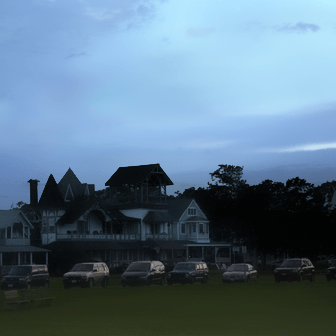}
 \includegraphics[width=\comparisonfigurewidth\linewidth]{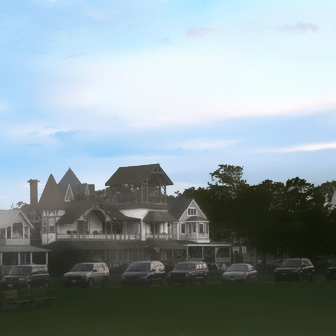}
 \includegraphics[width=\comparisonfigurewidth\linewidth]{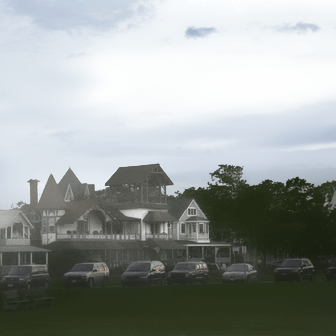}
 
 \includegraphics[width=\comparisonfigurewidth\linewidth]{comparisons_v2/original.png}
 \includegraphics[width=\comparisonfigurewidth\linewidth]{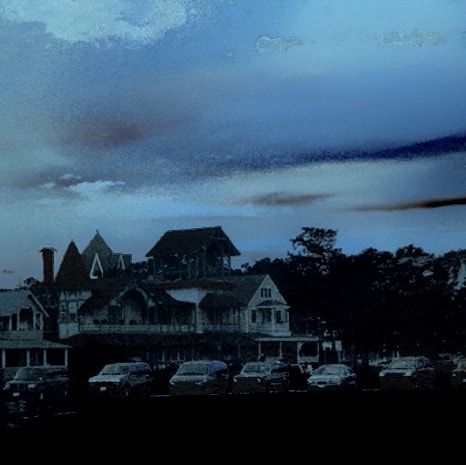}
 \includegraphics[width=\comparisonfigurewidth\linewidth]{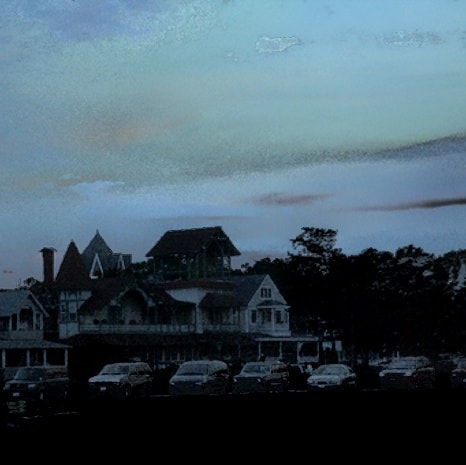}
 \includegraphics[width=\comparisonfigurewidth\linewidth]{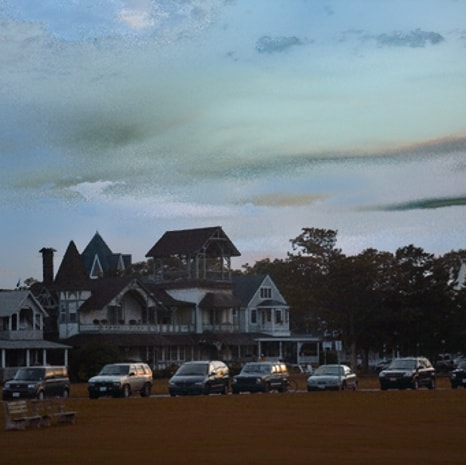}
 \includegraphics[width=\comparisonfigurewidth\linewidth]{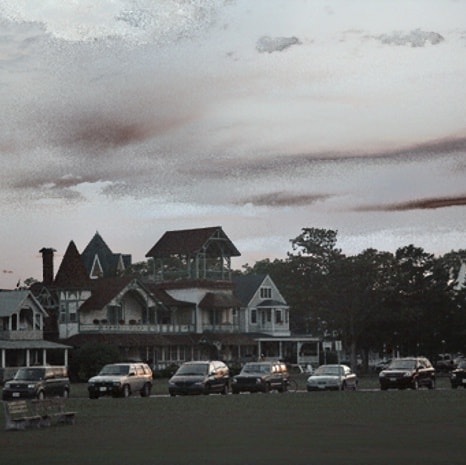}
 
 \includegraphics[width=\comparisonfigurewidth\linewidth]{comparisons_v2/original.png}
 \includegraphics[width=\comparisonfigurewidth\linewidth]{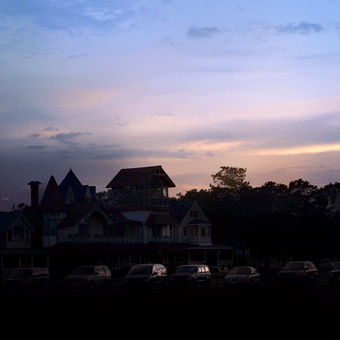}
 \includegraphics[width=\comparisonfigurewidth\linewidth]{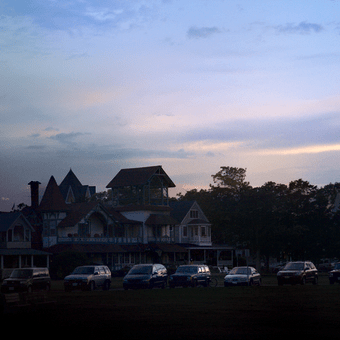}
 \includegraphics[width=\comparisonfigurewidth\linewidth]{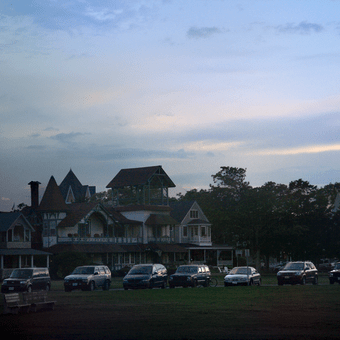}
 \includegraphics[width=\comparisonfigurewidth\linewidth]{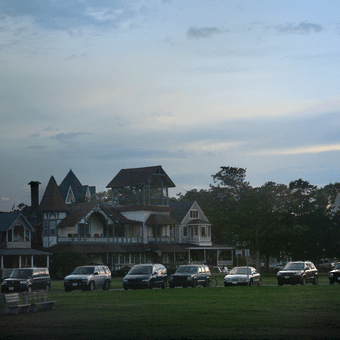}
\end{center}
\vspace{\figurevspace} 
\caption{Comparisons with existing methods. The same input image is shown on the left. From top to bottom, we show frames from a retrieved reference video, the results from~\cite{Li:2018:CFS}, \cite{Shih:2013:DHD}, and our method, respectively.
}
\label{fig:comparisons}
\end{figure}

\paragraph{Qualitative results.}

\Fref{fig:result_gallery} shows our results based on a variety of outdoor images from the MIT-Adobe 5K dataset~\cite{Bychkovsky:2011:MITAdobe5K}.
Our method can robustly handle input images with different semantic compositions and effectively synthesize illumination changes over time.
\Fref{fig:comparisons} shows qualitative comparisons between our results and those from the two baseline methods~\cite{Shih:2013:DHD,Li:2018:CFS}.
The input image is repeatedly shown in the first column in \Fref{fig:comparisons}.
The first row shows a set of frames from a retrieved reference video.
Starting from the second row, we show the results generated by~\cite{Li:2018:CFS},~\cite{Shih:2013:DHD}, and our method, respectively.
In many cases, both baseline methods produce unrealistic images, which is mainly because the scene in the reference video does not perfectly match the input image.
For both baseline methods, the color tone changes are driven by the retrieved reference video and may present noticeable visual artifacts if the reference video has semantic composition significantly different from the input image.
In contrast, our method can effectively generate plausible color tone changes over time without a reference video.
Also the color tone changes in our results are more visually pleasing due to the use of both AMOS and TLVDB datasets.

\paragraph{Computation time.}
At inference time, we only need $\generatortlvdb$ to synthesize time-lapse videos.
The inference of $\generatortlvdb$ takes about $0.02$ seconds with a GPU and $0.8$ seconds with a CPU.
The guided upsampling step takes about $0.1$ seconds on a CPU for an original input image of resolution $512 \times 512$.
In contrast, Shih~\etal's method~\cite{Shih:2013:DHD} takes $58$ seconds for a $700$-pixels width image on CPU, and Li~\etal's method~\cite{Li:2018:CFS} takes $6.3$ seconds for $768 \times 384$ resolution on GPU.
Therefore, our method is much faster than existing methods and is more suitable for deployment on mobile devices.

\subsection{Discussion}
\paragraph{Ablation studies.}

\begin{figure}[t]
\begin{center}
\centering
    \includegraphics[width=0.19\linewidth]{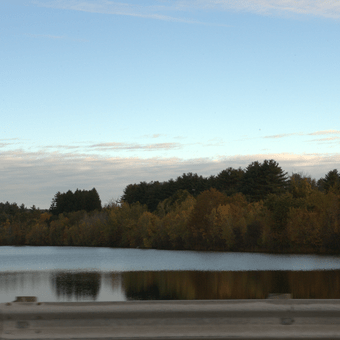}
    \includegraphics[width=0.19\linewidth]{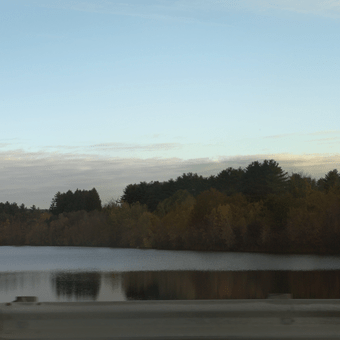}
    \includegraphics[width=0.19\linewidth]{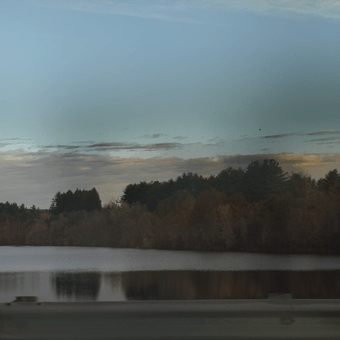}
    \includegraphics[width=0.19\linewidth]{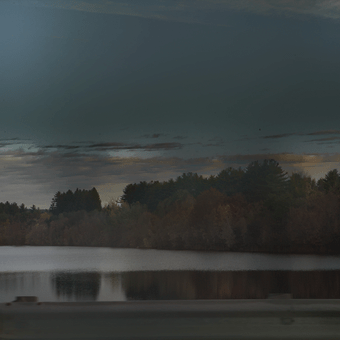}
    \includegraphics[width=0.19\linewidth]{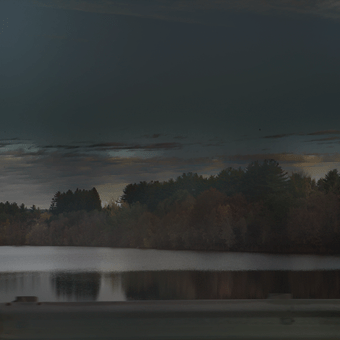}
    
    \includegraphics[width=0.19\linewidth]{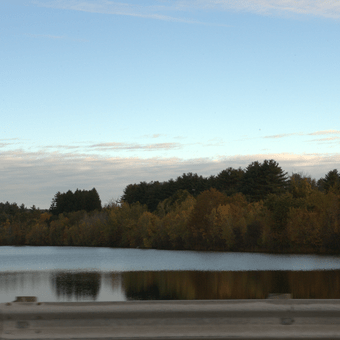}
    \includegraphics[width=0.19\linewidth]{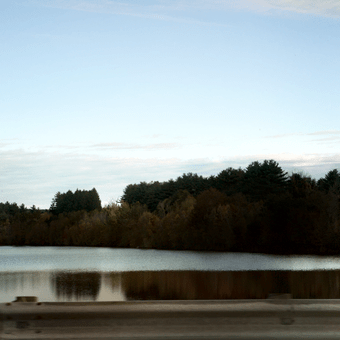}
    \includegraphics[width=0.19\linewidth]{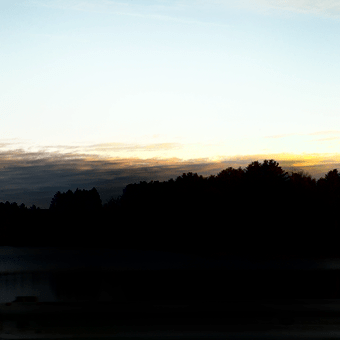}
    \includegraphics[width=0.19\linewidth]{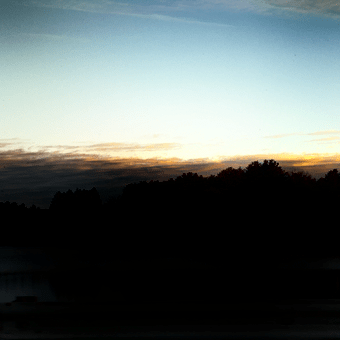}
    \includegraphics[width=0.19\linewidth]{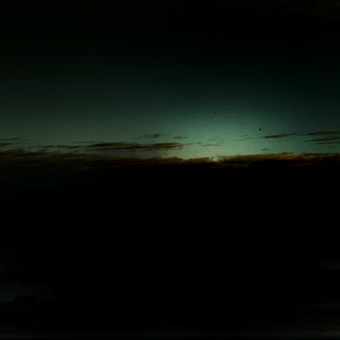}
    
    \includegraphics[width=0.19\linewidth]{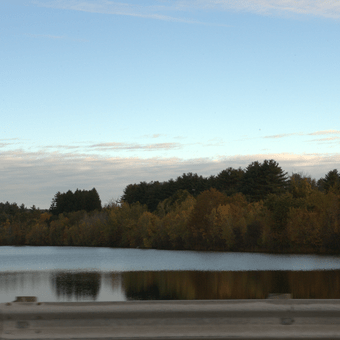}
    \includegraphics[width=0.19\linewidth]{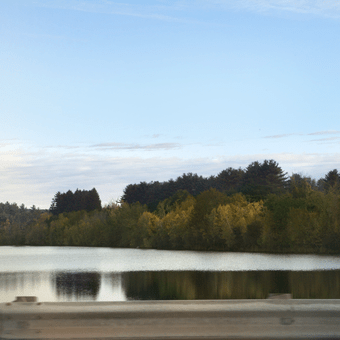}
    \includegraphics[width=0.19\linewidth]{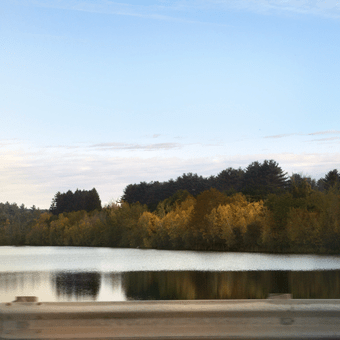}
    \includegraphics[width=0.19\linewidth]{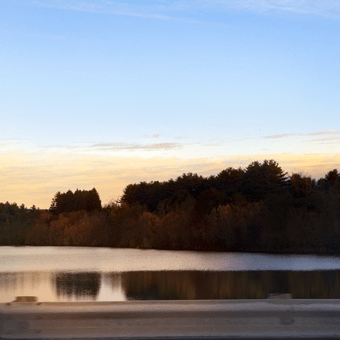}
    \includegraphics[width=0.19\linewidth]{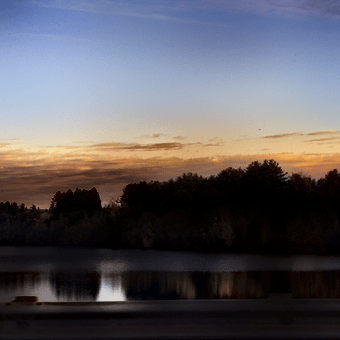}
\end{center}
\vspace{\figurevspace} 
\caption{Ablation study of our algorithm.
The same input is shown on the left.
From top to bottom, we show results from (A) a vanilla cGAN, (B) our multi-frame joint conditional GAN only, and (C) our full algorithm, respectively.
}
\label{fig:ablation_study}
\end{figure}

We conduct ablation studies to verify important components of our proposed method.
In \Fref{fig:ablation_study}, we show some qualitative results to compare with our own baselines: (A) a vanilla cGAN, (B) our multi-frame joint conditional GAN without the multi-domain training, and (C) our full algorithm with both the multi-frame joint conditional GAN and the multi-domain training.
Method (A) only changes the overall brightness without considering color tone changes, especially at a transition time such as the sunrise and the sunset.
This issue is because there can be various illumination changes at a specific time due to locations, season, and weather changes, which confuses the generator. 
Thus, the generator is likely to change brightness only as the easiest way to fool the discriminator.
Method (B) effectively captures illumination changes over time by considering the context of the entire sequence.
In many cases, however, it produces clipped pixels and less saturated color tone, because the AMOS dataset consists of footages captured by surveillance cameras which are visually less interesting.
Our full algorithm (C) overcomes this limitation by jointly learning from both the AMOS dataset and the TLVDB dataset.


\paragraph{Evaluation of multi-domain training.}
\Fref{fig:multi_domain_before_after} shows some examples after translating the output of $\generatortlvdb$ into the domain of AMOS dataset $\amos$ using $\generatoramos$ for the conditional training.
Directly training $\generatortlvdb$ using $\discriminatoramos$ fails due to the domain discrepancy of the two datasets such as different color tones and scene compositions.
As shown in the figure, $\generatoramos$ can effectively change the output of $\generatortlvdb$ to fool $\discriminatoramos$ and get conditional training signals from it.

\newcommand\multidomainfigurewidth{0.19}
\begin{figure}[t]
\begin{center}
\centering

    \includegraphics[width=\multidomainfigurewidth\linewidth]{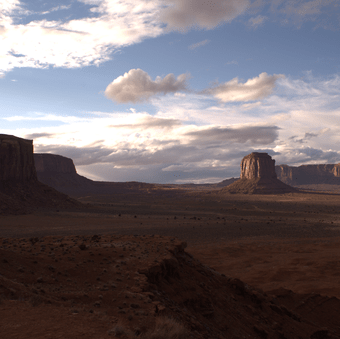}
    \includegraphics[width=\multidomainfigurewidth\linewidth]{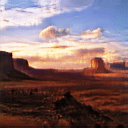}
    \includegraphics[width=\multidomainfigurewidth\linewidth]{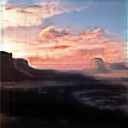}
    \includegraphics[width=\multidomainfigurewidth\linewidth]{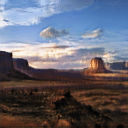}
    \includegraphics[width=\multidomainfigurewidth\linewidth]{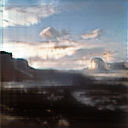}
    
    \includegraphics[width=\multidomainfigurewidth\linewidth]{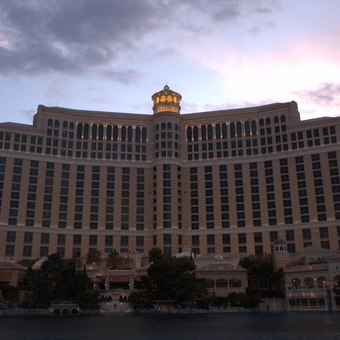}
    \includegraphics[width=\multidomainfigurewidth\linewidth]{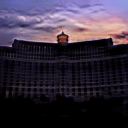}
    \includegraphics[width=\multidomainfigurewidth\linewidth]{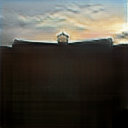}
    \includegraphics[width=\multidomainfigurewidth\linewidth]{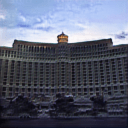}
    \includegraphics[width=\multidomainfigurewidth\linewidth]{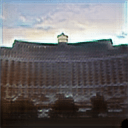}
\end{center}
\vspace{\figurevspace} 
\caption{The effect of using $\generatoramos$. The input image is shown on the left. 
Two image pairs of before (the 2nd and 4th columns) and after $\generatoramos$ (the 3rd and 5th columns) are shown to verify the multi-domain training.
}
\label{fig:multi_domain_before_after}
\end{figure}

\section{Conclusions}
\label{sec:conclusions}

In this paper, we presented a novel framework for the time-lapse video synthesis from a single outdoor image.
Given an input image, our conditional generative adversarial network can predict the illumination changes over time by using the timestamp as the control variable.
Compared to other methods, we do not require semantic segmentation or a reference video to guide the generation of the output video.

\begin{wrapfigure}{r}{0.42\linewidth}
\vspace{-5mm}
\begin{center}
    \includegraphics[width=0.48\linewidth]{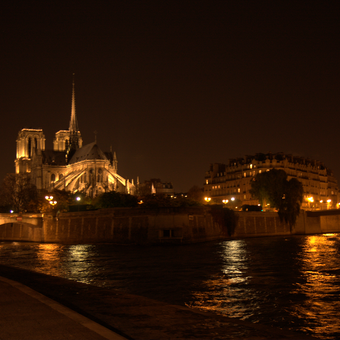}
    \includegraphics[width=0.48\linewidth]{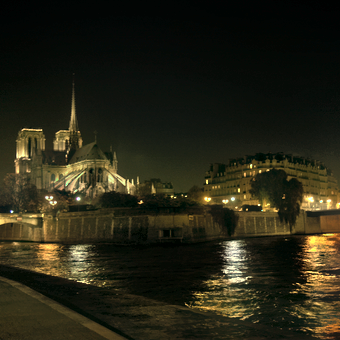}
    \includegraphics[width=0.48\linewidth]{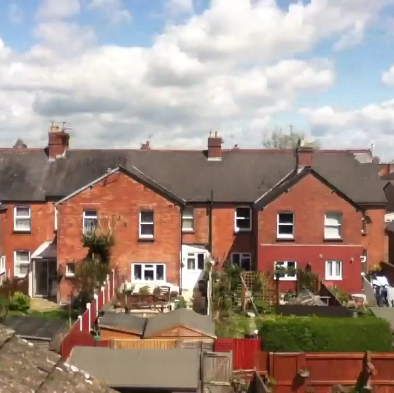}
    \includegraphics[width=0.48\linewidth]{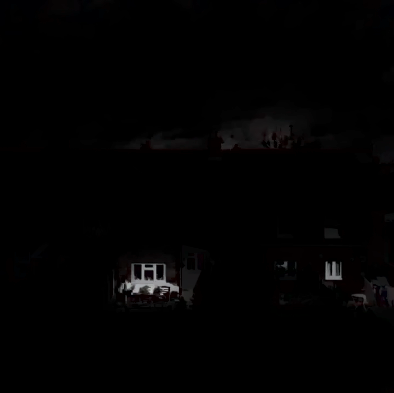}
\end{center}
\vspace{-5mm}
\caption{Failure cases.}
\label{fig:failure_cases}
\vspace{-3mm}
\end{wrapfigure}

Our method still has some limitations.
As shown in~\Fref{fig:failure_cases}, our method fails to hallucinate daytime images from a nighttime input where most parts of the input are very dark.
In some cases, our method fails to generate artificial lighting in regions such as building windows.
In addition, our method only changes the color tones of a given input image without introducing any motions such as moving objects.
It would be interesting to combine our approach with frame prediction or motion synthesis techniques~\cite{Xiong:2018:MDGAN} to generate time-lapse videos with both interesting motions and illumination changes.
We also plan to extend our approach to support additional semantic controls such as sunrise and sunset times in the prediction results~\cite{Karacan:2016:AGAN}.
Finally, we would like to investigate using our synthesis framework with an implicit control variable for general video synthesis tasks.

\paragraph{\small{Acknowledgements.}}
\small{
This work was supported by National Research
Foundation of Korea (NRF) grant funded by the Korea government (MSIP) (NRF-2016R1A2B4014610) and Institute of Information \& Communications Technology Planning \& Evaluation (IITP) grant funded by the Korea government (MSIT) (2014-0-00059). Seonghyeon Nam was partially supported by Global Ph.D. Fellowship Program through the National Research Foundation of Korea (NRF) funded by the Ministry of Education (NRF2015H1A2A1033924).
}

{\small
\bibliographystyle{ieee}
\bibliography{paper}
}

\end{document}